\documentclass[conference]{IEEEtran}
\IEEEoverridecommandlockouts

\usepackage[numbers,sort&compress]{natbib}
\usepackage{amsmath,amssymb,amsfonts}
\usepackage{algorithmic}
\usepackage{graphicx}
\usepackage{textcomp}
\usepackage{xcolor}
\usepackage{booktabs}
\usepackage{xurl}
\def\BibTeX{{\rm B\kern-.05em{\sc i\kern-.025em b}\kern-.08em
    T\kern-.1667em\lower.7ex\hbox{E}\kern-.125emX}}
\usepackage{hyperref}
\hypersetup{
    colorlinks   = true,
    citecolor    = blue,
    urlcolor    =  blue
}
\usepackage{makecell}
\usepackage{tablefootnote}

\newcommand{\hf}[2]{\raisebox{-2.2pt}{\includegraphics[scale=0.09]{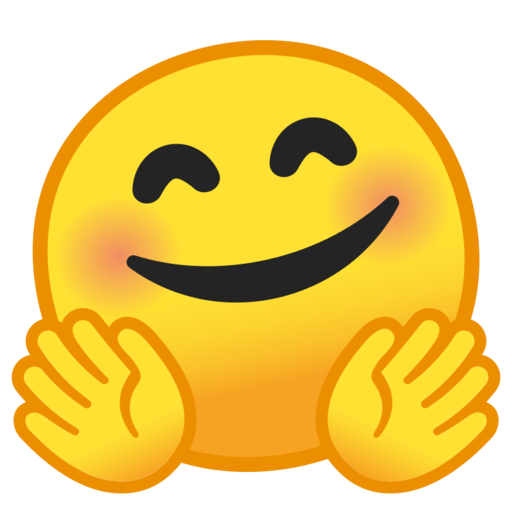}}~\href{#1}{\texttt{#2}}}

\newcommand{\gh}[2]{\raisebox{-2.2pt}{\includegraphics[scale=0.02]{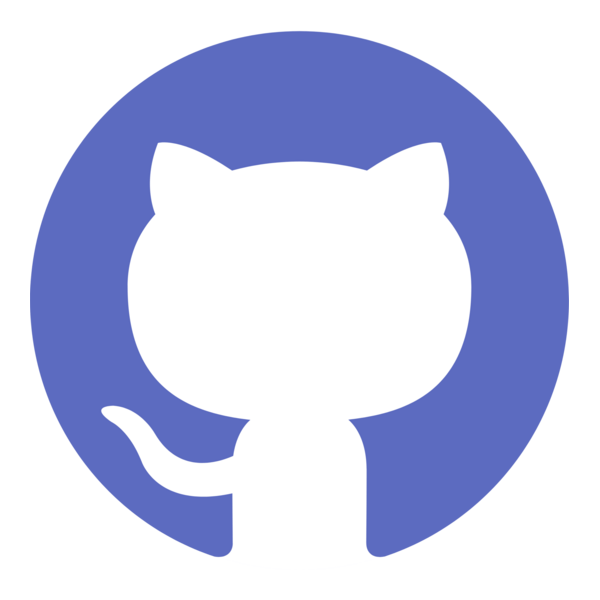}}~\href{#1}{\texttt{#2}}}

\begin{document}

% Group 10
\title{Environmental Drivers of Respiratory Disease: A District Level Analysis}

\author{
\IEEEauthorblockN{%
Rahim Iqbal,
Asfi Ahamed,
Izzath Nisfer,
Shazan Shaheed,
Muhammadu Ilham,\\
Nathali Athukorala,
Madara Mendis,
Nisansa de Silva,
Sandareka Wickramanayake}
\IEEEauthorblockA{Dept.\ of Computer Science \& Engineering, University of Moratuwa, Sri Lanka.\\
\texttt{\{rahimi.23, asfia.23, izzathn.23, shazans.23, muhammadui.23,}\\
\texttt{nathalia, madara.21, NisansaDds, sandarekaw\}@cse.mrt.ac.lk}
}
}

\maketitle

\begin{abstract}

Sri Lanka has experienced a decade of progressive forest degradation and rising atmospheric pollution, yet district-level respiratory admissions have paradoxically declined, pointing to the confounding role of healthcare access. This study addresses that gap by constructing an 11-year (2014- 2024) panel dataset across all 25 administrative districts, integrating satellite-derived vegetation indices, fire radiative power, pollutant concentrations (particulate matter (PM\textsubscript{2.5}), nitrogen dioxide (NO\textsubscript{2}), sulfur dioxide (SO\textsubscript{2})), carbon flux metrics and population- normalized respiratory admission rates. Two temporally validated XGBoost models were created for annual district-level respiratory rate ($R^2 = 0.937$) and monthly PM\textsubscript{2.5} concentration. ($R^2 = 0.976$) with generalization validated in 21 out of 25 districts (Mean Absolute Percentage Error (MAPE\,$\leq 20\%$)). Shapley Additive Explanations (SHAP) analysis identified cumulative air quality burden as the strongest predictor of respiratory rate variance in the model (80.1\% of the aggregate environmental SHAP signal), ahead of forest degradation (15.6\%) and fire activity (4.3\%). The Forest-Air-Health (FAH) Risk Index used these SHAP-derived weights to find the districts with the highest risk: Colombo (FAH\,$ = 0.802$), Gampaha (0.708), and Kalutara (0.682). These findings present the inaugural evidence-based, district-level framework correlating environmental degradation with respiratory health in Sri Lanka, establishing a quantitative basis for focused public health and environmental policy.
\end{abstract}

\begin{IEEEkeywords}
forest degradation, air quality, respiratory health, XGBoost, SHAP, FAH Risk Index, district-level analysis, Sri Lanka
\end{IEEEkeywords}

%----------------------------------------------------------------------
\section{Introduction}
%----------------------------------------------------------------------
Sri Lanka lost about 230,000 hectares of tree cover~\cite{hansen2013high} from 2001 to 2024. This was due to biomass burning, agricultural fires, and urbanization~\cite{Asia2023Time} in the North Western, North Central, and Northern provinces. Long-term exposure to PM\textsubscript{2.5}, NO\textsubscript{2}, and SO\textsubscript{2} is linked to bronchitis, asthma, Chronic Obstructive Pulmonary Disease (COPD), and respiratory mortality~\cite{who2024Global,priyankara2021ambient}. However, the environmental factors that affect respiratory admissions in Sri Lanka's 25 districts have not yet been measured at the district level~\cite{Ministry2022Annual}.

Previous research reveals three significant deficiencies: remote sensing analyses conclude at land cover classification without establishing connections to health outcomes~\cite{vijitharan2022assessment}; air quality forecasting has been limited to Colombo~\cite{rathnayaka2024development}; and epidemiological studies focus on individual pollutants over short time frames~\cite{priyankara2021ambient}. This study fills these gaps by creating an 11-year harmonized panel dataset that includes forest, fire, atmospheric, and health variables for all 25 districts. It also trains temporal XGBoost models~\cite{chen2016xgboost} to predict district-level respiratory rates and monthly PM\textsubscript{2.5}; uses SHAP-based interpretation to measure environmental drivers; and creates the Forest-Air-Health (FAH) Risk Index to rank all 25 districts by environmental health risk using weights informed by SHAP importance. \hf{https://huggingface.co/datasets/shazan18/environmental-drivers-respiratory-disease-sri-lanka}{Data} and \gh{https://github.com/asfiahamed0404/Environmental-Drivers-of-Respiratory-Disease-Sri-Lanka}{code} for this work are publicly available.
%----------------------------------------------------------------------
\section{Related Work}
%----------------------------------------------------------------------
Hansen et al.~\cite{hansen2013high} offer the fundamental 30\,m global forest cover change product that supports analyses of tropical deforestation. The Northern Province of Sri Lanka experienced the most significant tree cover loss compared to 2010 levels~\cite{World2024Global,ranasinhe2024deforestation}. Additionally, Vijitharan et al.~\cite{vijitharan2022assessment} illustrated canopy fragmentation in Vavuniya utilizing Google Earth Engine; however, neither study established a connection between the findings and health outcomes. NASA MERRA-2~\cite{NASA2024Giovanni} and CAMS EAC4~\cite{inness2019cams} reanalysis products provide monthly PM\textsubscript{2.5}, SO\textsubscript{2}, and NO\textsubscript{2} at a coarse spatial resolution. XGBoost has become the leading algorithm for air quality regression due to its ability to capture non-linear feature interactions~\cite{chen2016xgboost}. Rathnayaka et al.~\cite{rathnayaka2024development} attained commendable Air Quality Index (AQI) forecasting accuracy in Colombo, albeit limited to a single district, while Priyankara et al.~\cite{priyankara2021ambient} discerned short-term PM\textsubscript{2.5} correlations with respiratory admissions in Kandy. World Health Organization (WHO) data substantiate significant pollution-related Disability-Adjusted Life Years (DALYs) at the national level~\cite{who2024Global,who2022Ambient}; however, no previous study has concurrently evaluated various pollutants, deforestation, and respiratory disease burden across all 25 districts of Sri Lanka.
%----------------------------------------------------------------------
\section{Methodology}
%----------------------------------------------------------------------

\subsection{Data Sources and Panel Construction}
The panel dataset covers the years 2014 to 2024, with 3,300 observations (25 districts $\times$ 12 months $\times$ 11 years). We made district composites by putting point measurements (active fire locations from SUOMI VIIRS C2) and raster measurements (MERRA-2 atmospheric variables, VIIRS 8-day vegetation indices) on top of HDX district boundary polygons. Daily and 8-day satellite data were averaged to make monthly composites. Annual health records were spread out evenly over the months of the year. After fire-NaN imputation, the final dataset had 3,300 observations and 115 feature-engineered variables that looked at deforestation, fire activity, air pollution, vegetation indices, demographics and health outcomes.

\begin{table*}[htbp]
\caption{Data Sources and Feature Domains}
\label{tab:datasources}
\begin{center}
\begin{tabular}{lll}
\hline
\textbf{Domain} & \textbf{Source} & \textbf{Key Variables} \\
\hline
Forest Cover \& Carbon & GFW (threshold 30\%) \cite{World2024Global}, Hansen 
GFC \cite{hansen2013high} & tc\_loss\_ha, ExtentIn2010, carbon stocks, 
emissions, net flux \\
%\hline
Vegetation & VIIRS 8-day via GLAM \cite{GLAM2024VIIRS} & VIM, VIM anomaly, 
VIM climatology, VIM range \\
%\hline
Fire Activity & NASA FIRMS, SUOMI VIIRS C2 \cite{FIRMS2024SUOMI} & Fire Radiative Power (FRP) 
mean/total, brightness, fire type \\
%\hline
PM\textsubscript{2.5}, SO\textsubscript{2} & MERRA-2 via Giovanni 
\cite{NASA2024Giovanni} & PM\textsubscript{2.5} $\mu$g/m$^3$, SO\textsubscript{2} 
$\mu$g/m$^3$ \\
%\hline
NO\textsubscript{2} & CAMS EAC4 reanalysis \cite{inness2019cams} & 
NO\textsubscript{2} $\mu$g/m$^3$ \\
%\hline
Health & MoH Annual Health Bulletin \cite{Ministry2022Annual} & \makecell[l]{Bronchitis/COPD 
(J40-J44), Asthma (J45-J46) admissions,\\deaths, Case Fatality Rate (CFR)} \\
%\hline
Population & DCS, Sri Lanka \cite{Department2024Mid} & Total, male, female 
mid-year population (1k) \\
%\hline
Boundaries & HDX Admin Level 2 \cite{OCHA2025Lanka} & District polygons \\
\hline
\end{tabular}
\end{center}
\end{table*}

\subsection{Spatial Processing of Atmospheric Data}
District-level PM\textsubscript{2.5} and SO\textsubscript{2} 
concentrations were derived from MERRA-2 M2TMNXAER v5.12.4~\cite{NASA2024Giovanni} gridded 
data ($0.5^\circ \times 0.625^\circ$ resolution) using spatial 
area-weighted aggregation. Each grid cell centroid was converted to 
a rectangular polygon using its half-widths ($\pm0.25^\circ$ 
latitude, $\pm0.3125^\circ$ longitude) and intersected with Sri 
Lanka's district boundaries. The area-weighted concentration $C_d$ 
for district $d$ is:

\begin{equation}
C_d = \frac{\sum_i C_i \cdot A_{i \cap d}}{\sum_i A_{i \cap d}}
\label{eq:areaweight}
\end{equation}

where $C_i$ is the amount of grid cell $i$ and $A_{i \cap d}$ is the area where grid cell $i$ and district $d$ overlap. This area-weighting method gets rid of spatial sampling bias that can happen when you use point-based or nearest-neighbor assignments across uneven administrative boundaries. To change kg\,m$^{-3}$ to $\mu$g\,m$^{-3}$, you had to multiply by $10^9$. We got the NO2 concentrations from the CAMS EAC4 monthly reanalysis~\cite{inness2019cams} (1000\,hPa level) and changed them from mass mixing ratios (kg\,kg$^{-1}$) to $\mu$g\,m$^{-3}$ by multiplying by air density (1.225\,kg\,m$^{-3}$) and $10^9$. Using the same area-weighted overlay in GeoPandas (EPSG:32644, UTM Zone 44N), we got district-level values. This made sure that the method used for PM\textsubscript{2.5} and SO\textsubscript{2} processing was the same.

\subsection{Data Preprocessing}
\label{sec:preprocessing}
There were six preprocessing steps: (i) normalizing column names to get rid of whitespace and BOM characters; (ii) filling in missing values for fire data by setting FRP to 0 and fire type to "No fire" for district-months with no hot-spots found by SUOMI VIIRS C2; (iii) aligning the data over time by spreading annual health records evenly across 12 months, which was a necessary simplification that the yearly respiratory model took care of; (iv) labeling and cyclic encoding of province, district, fire type, and month variables; (v) feature engineering produced 75 new variables across demographic, health, forest, fire, vegetation, and composite domains, including temporal lags (one-month and three-month), rolling three-month averages, and year-over-year differences for seven key variables (PM\textsubscript{2.5}, SO\textsubscript{2}, NO\textsubscript{2}, respiratory rate, VIM, FRP and tree cover loss) computed in district-ordered fashion to avoid data leakage; and (vi) quality validation confirming 3,300 complete records with no missing values outside lag-derived features.

\subsection{Exploratory Data Analysis Pipeline}
The Exploratory Data Analysis (EDA) pipeline consisted of four stages: (i) distribution histograms with Kernel Density Estimation (KDE) overlays to evaluate marginal distributions and skewness for 14 key variables; (ii) Pearson and Spearman cross-correlation matrices and heatmaps to detect inter-domain dependencies among pollution, vegetation, fires, and health, including a cumulative pollution index that integrates scaled PM\textsubscript{2.5}, SO\textsubscript{2}, and NO\textsubscript{2}; (iii) year-on-year national aggregations with linear slope coefficients to analyze temporal trends; and (iv) Principal Component Analysis (PCA) for dimensionality reduction and $K$-means clustering, with the number of clusters $k = 4$ determined through the elbow method to identify district archetypes.

\subsection{Machine Learning Pipeline}
Two temporal XGBoost regressors were developed using a strict
chronological protocol: training years 2015--2020, testing years
2021--2024, and 2014 excluded due to insufficient lagged features.
There were no separate calendar validation years; instead, validation
for model selection was performed via time-series cross-validation
folds drawn entirely from within the 2015--2020 training window, so
the 2021--2024 test years were never used for fitting or tuning.

\textbf{Model 1: Yearly Respiratory Rate (Primary).} 
This model works at an annual resolution to avoid artifacts that come from spreading out annual health records over months. The dataset consisted of 250 district-year observations (25 districts 
$\times$ 10 years), divided into 150 training observations and 100 test observations. The 49-variable feature vector comprised 42 numerical environmental variables, along with categorical variables for calendar year, district, and province. It also included cross-year auto regressive health lags (previous year, two years prior, two-year mean, and year-over-year change) to capture temporal dependencies without data leakage.

\textbf{Model 2: Monthly PM\textsubscript{2.5} Forecasting.} 
This model works with monthly data and has a 43-variable feature vector that includes vegetation indices and anomalies, forest cover, biomass, FRP, brightness, smoke proxy, co-pollutant lags and rolling averages, population density, cyclic month encoding, and district and season as categorical variables.

Both models were tuned by randomized search over ensemble size, learning rate, maximum depth, sub-sampling ratios, and L1/L2 regularization, with time-series cross-validation (3 folds for the yearly model; 4 folds for the monthly model). The search ranges, final selected values, and the fixed random seed are reported in Table~\ref{tab:hyperparams} for reproducibility. A temporal rather than random split was used to test whether environmental covariates generalize to future unseen years. SHAP TreeExplainer was used to decompose predictions into feature-level contributions, aggregated across three environmental domains (Forest, Fire, and Air Quality) to obtain the SHAP-informed weights used in the FAH Risk Index.

\begin{table}[htbp]
\caption{XGBoost Hyperparameter Search Space and Final Values}
\label{tab:hyperparams}
\begin{center}
\small
\begin{tabular}{lccc}
\hline
\textbf{Hyperparameter} & \textbf{Search Range} &
\textbf{Yearly} & \textbf{Monthly} \\
\hline
n\_estimators       & \{300,500,700,900\}     & 300  & 900  \\
learning\_rate      & \{0.01,0.03,0.05,0.08\} & 0.05 & 0.08 \\
max\_depth          & \{3,4,5,6\}             & 3    & 3    \\
subsample           & \{0.7,0.8,0.9,1.0\}     & 0.7  & 1.0  \\
colsample\_bytree   & \{0.7,0.8,0.9,1.0\}     & 0.9  & 0.7  \\
min\_child\_weight  & \{1,3,5\}               & 5    & 1    \\
reg\_alpha (L1)     & \{0,0.1,0.3\}           & 0    & 0.1  \\
reg\_lambda (L2)    & \{1,1.5,2\}             & 1    & 1.5  \\
\hline
\multicolumn{4}{l}{Random seed $=$ 42, fixed across all
stochastic components.}\\
\end{tabular}
\end{center}
\end{table}

%----------------------------------------------------------------------
\section{Results and Discussion}
%----------------------------------------------------------------------

\subsection{Descriptive Statistics of the Dataset}
The dataset contains 3,300 observations from 25 districts in 9 provinces over 11 years (2014-2024) at a rate of 300 records per year, with 115 engineered features covering forest degradation, fire activity, air quality, vegetation health and respiratory health outcomes.

\textbf{Air quality.} The mean concentration of PM\textsubscript{2.5} was 16.51\,$\mu$g/m$^3$ (SD\,$= 4.57$) and above the WHO guideline of 5\,$\mu$g/m$^3$ for all observations. The Northern (20.02), North Western (17.93) and Western (17.11\,$\mu$g/m$^3$) provinces recorded the highest pollution levels. Concentrations peaked between May and July due to biomass burning and were lowest in October (13.47\,$\mu$g/m$^3$). Average SO\textsubscript{2} and NO\textsubscript{2} were 1.77 and 7.64 $\mu$g/m$^3$, respectively.

\textbf{Degradation of forests.} Mean tree cover loss was 473\,ha (SD\,$= 419$) with the highest losses in the North Western (970\,ha) and North Central (847\,ha) provinces. Although the carbon emissions were high (193,358 units), the net carbon flux was still negative (mean = -437,704), meaning the country was still a net carbon sink during the study period.

\textbf{Fire activity.}  Active fires were recorded in 70\% of observations ($n = 2{,}319$), with peak fire activity in August, when 89.7\% of district-months reported fire. Fire radiative power was highly right-skewed (mean\,$= 3.29$, median\,$= 2.81$) consistent with short, intense burning episodes.

\textbf{Respiratory health.} The total respiratory rate was between 0.007 and 21.45 per 1,000 (mean = 9.17), with asthma (7.39 per 1,000) more common than bronchitis (1.78 per 1,000). District populations ranged from 95,000 to 2,480,000.

\subsection{Findings From Exploratory Data Analysis}

\textit{1) Cross-Domain Relationships:} The negative correlation of vegetation cover (VIM) with PM\textsubscript{2.5} ($r = -0.40$) confirms the vegetative filtering effect. The strong positive correlation between NO\textsubscript{2} and SO\textsubscript{2} ($r = 0.51$) suggests common emission sources. 
The composite pollution index is paradoxically positively correlated with both forest cover ($r = 0.34$) and tree cover loss ($r = 0.44$), implying that biomass burning in forested areas compensates for vegetative filtration. Forest cover and gross carbon emissions are strongly correlated ($r = 0.89$).

\textit{2) Temporal Trends:} At the national level, 
NO\textsubscript{2} and PM\textsubscript{2.5} have risen (slopes 
of $+0.15$ and $+0.06\,\mu$g/m$^3$/year, respectively), whereas 
tree cover loss (slope\,$= -130$\,ha/year), respiratory hospital 
admissions (slope\,$= -0.34$ per 1{,}000/year), and carbon 
emissions ($-$27{,}198\,Mg\,CO$_2$e/year) have declined. The fall 
in admissions is more likely attributable to improvements in 
healthcare access than to reduced pollution exposure, and lower 
carbon emissions are consistent with reduced deforestation intensity 
in recent years.

\textit{3) Deforestation and Respiratory Burden:} At the district 
level, deforestation alone is a poor predictor of respiratory 
outcomes. Jaffna and Colombo share nearly identical tree cover loss 
rates (0.8\% and 0.7\%) yet differ substantially in respiratory 
rates (13 vs.\ 6.2 per 1{,}000), a disparity likely attributable 
to Colombo's stronger healthcare infrastructure. Hambantota's 
elevated respiratory rates ($>$13 per 1{,}000) despite moderate 
deforestation further confirm that multiple exposure
pathways, not deforestation alone, are associated with respiratory outcomes.

\begin{figure*}[htbp]
\centerline{\includegraphics[width=\linewidth]{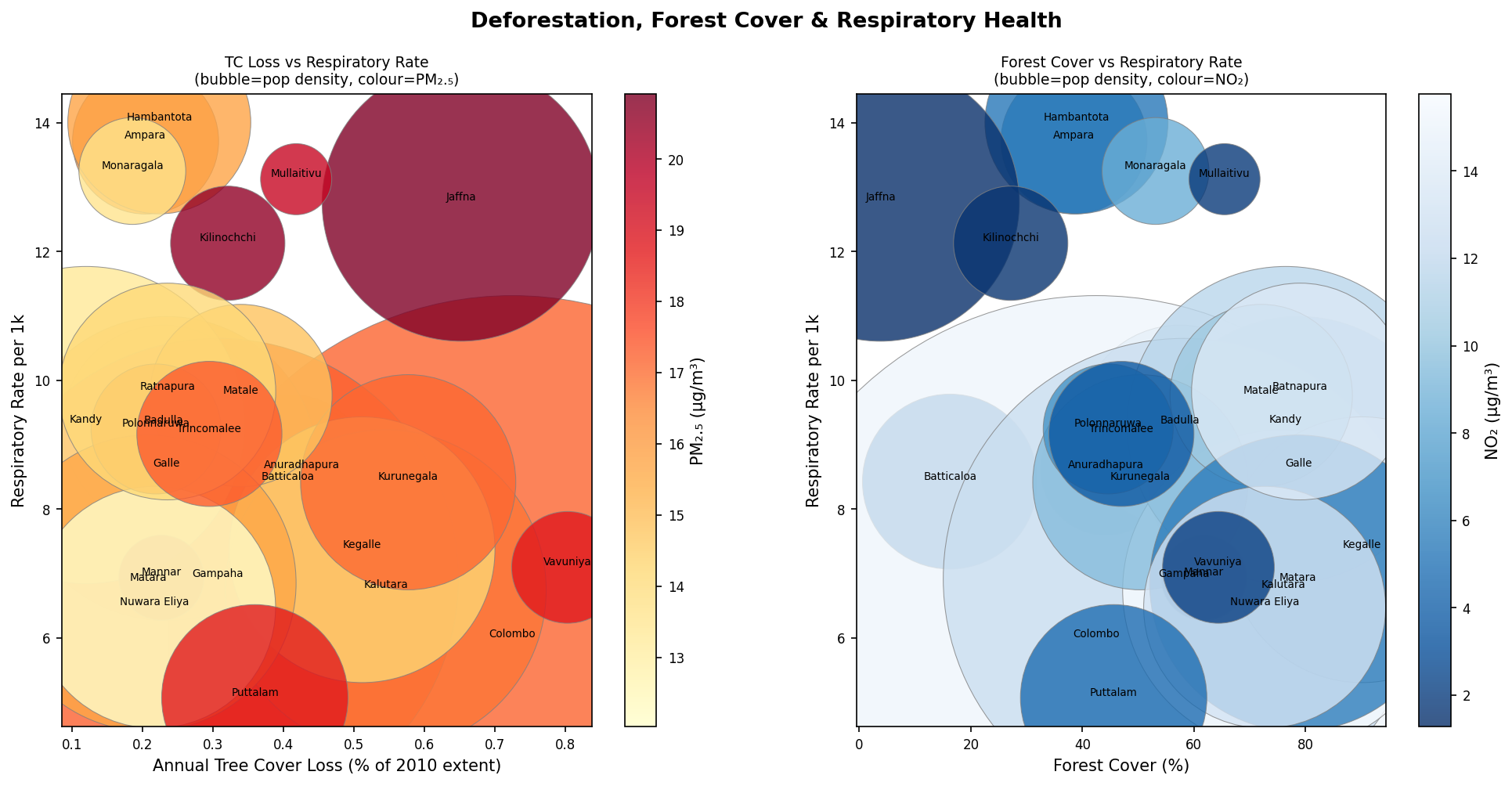}}
\caption{Deforestation, forest cover, and respiratory health across 
25 districts. Left: tree cover loss vs.\ respiratory rate 
(bubble\,$=$\,population density, color\,$=$\,PM\textsubscript{2.5}). 
Right: forest cover vs.\ respiratory rate 
(bubble\,$=$\,population density, color\,$=$\,NO\textsubscript{2}).}
\label{fig:bubble}
\end{figure*}

\begin{figure*}[htbp]
\centerline{\includegraphics[width=\linewidth]{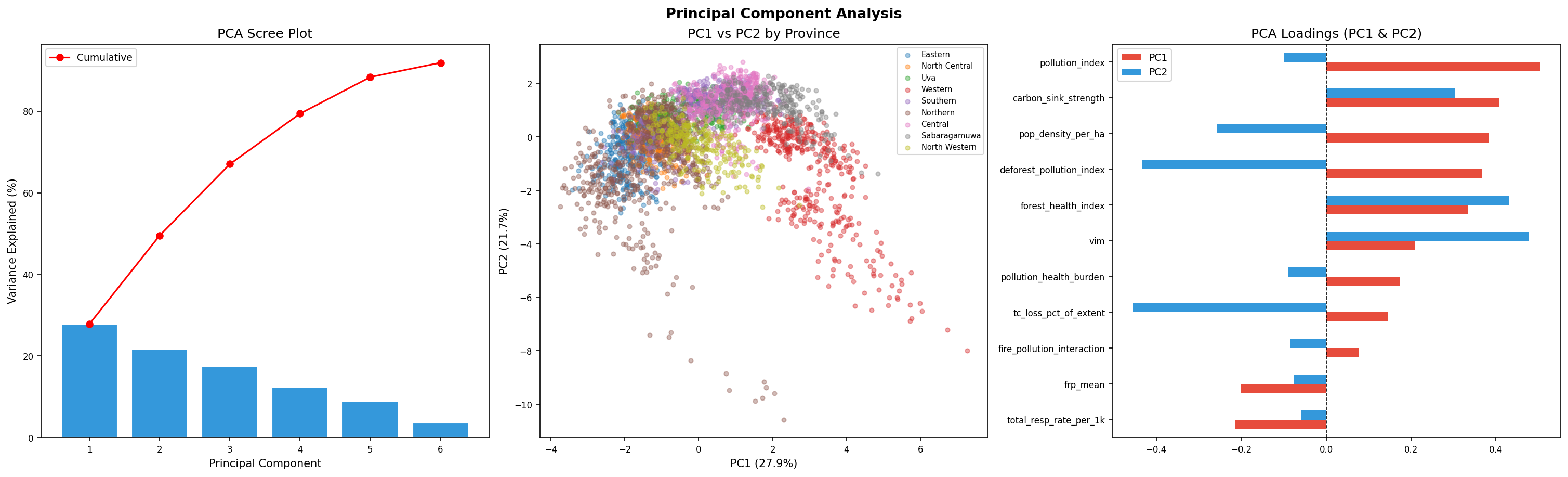}}
\caption{Principal Component Analysis. Left: scree plot showing six 
components capturing 95\% of variance. Center: PC1 vs.\ PC2 scatter 
colored by province, showing spatial clustering. Right: PCA 
loadings identifying pollution, forest health, and population as 
dominant dimensions.}
\label{fig:pca}
\end{figure*}

\textit{4.) Principal Component Analysis:} PCA showed six components accounting for 95.0\% of the total variance (Fig.~\ref{fig:pca}). PC1 (27.9\%) was involved in pollution intensity and carbon sink effectiveness. PC2 (21.7\%) was involved in forest health and vegetation status. Districts in the Northern Province clustered towards high PC1 and low PC2, indicating higher degradation and pollution, while the other provinces clustered towards low PC1 and high PC2, indicating comparatively healthy environmental conditions.

\begin{figure*}[htbp]
\centerline{\includegraphics[width=\linewidth]{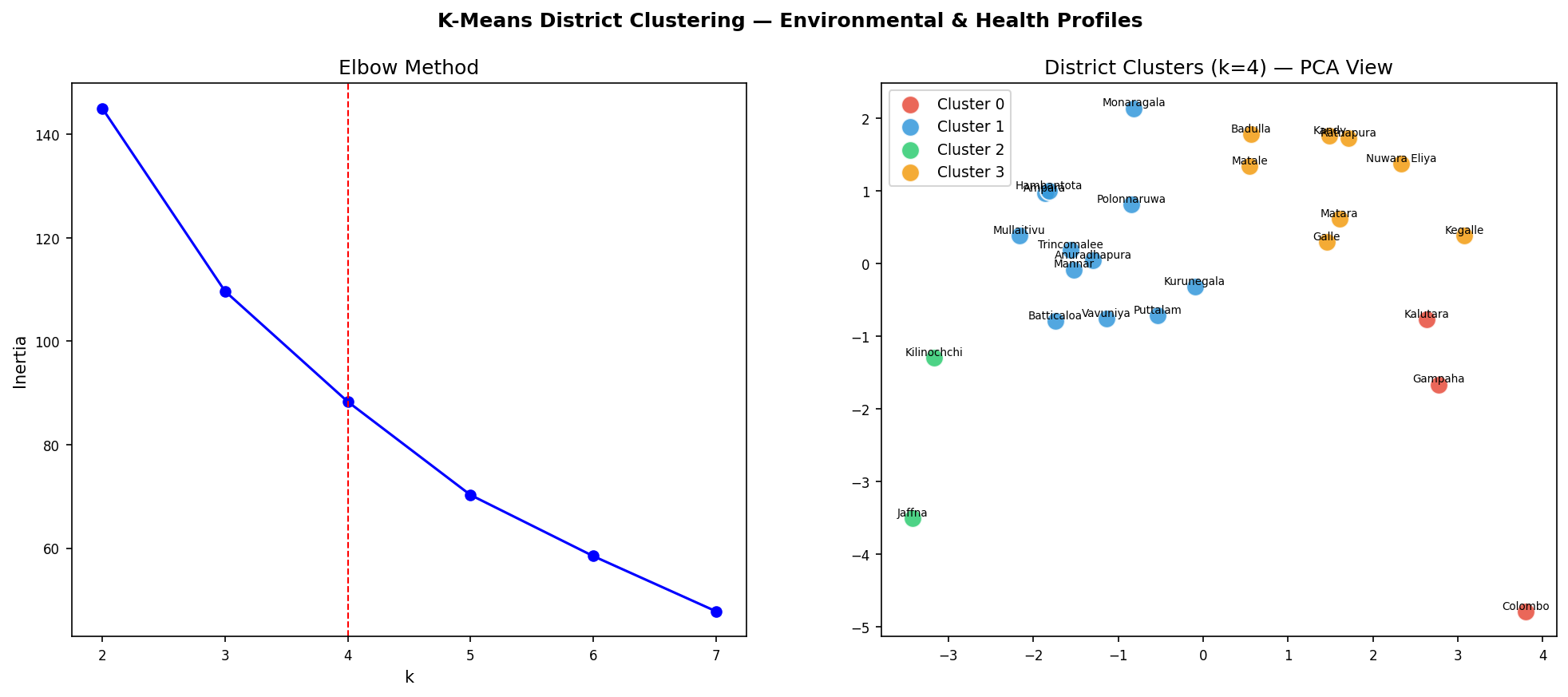}}
\caption{$K$-means district clustering ($k = 4$) based on nine 
environmental and health variables. Left: elbow plot for cluster 
selection. Right: PCA-projected cluster visualization with district 
labels.}
\label{fig:cluster}
\end{figure*}

\textit{5) District Clustering:} The $K$-means clustering ($k = 4$) grouped districts into four archetypes (Fig.~\ref{fig:cluster}): (i) high urbanization and pollution (Western Province); (ii) moderate pollution with high fire activity (North Central and North Western provinces); (iii) high forest cover with low pollution (Sabaragamuwa and Central provinces); and (iv) post-conflict 
Northern districts with high deforestation and mixed health outcomes.

\subsection{Model Performance}
Table~\ref{tab:model_perf} summarizes the evaluation metrics for 
both XGBoost models.

\begin{table}[htbp]
\caption{XGBoost Model Performance}
\label{tab:model_perf}
\begin{center}
\begin{tabular}{lcccc}
\hline
\textbf{Model} & \textbf{Level} & \textbf{Test $R^2$} & 
\textbf{MAE} & \textbf{CV $R^2$}\\
\hline
Respiratory Rate & Yearly & 0.937 & 0.776 & 0.797 \\
PM\textsubscript{2.5} ($\mu$g/m$^3$) & Monthly & 0.976 & 0.520 
& 0.943 \\
\hline
\end{tabular}
\end{center}
\end{table}

\begin{figure*}[htbp]
\centerline{\includegraphics[width=\linewidth]{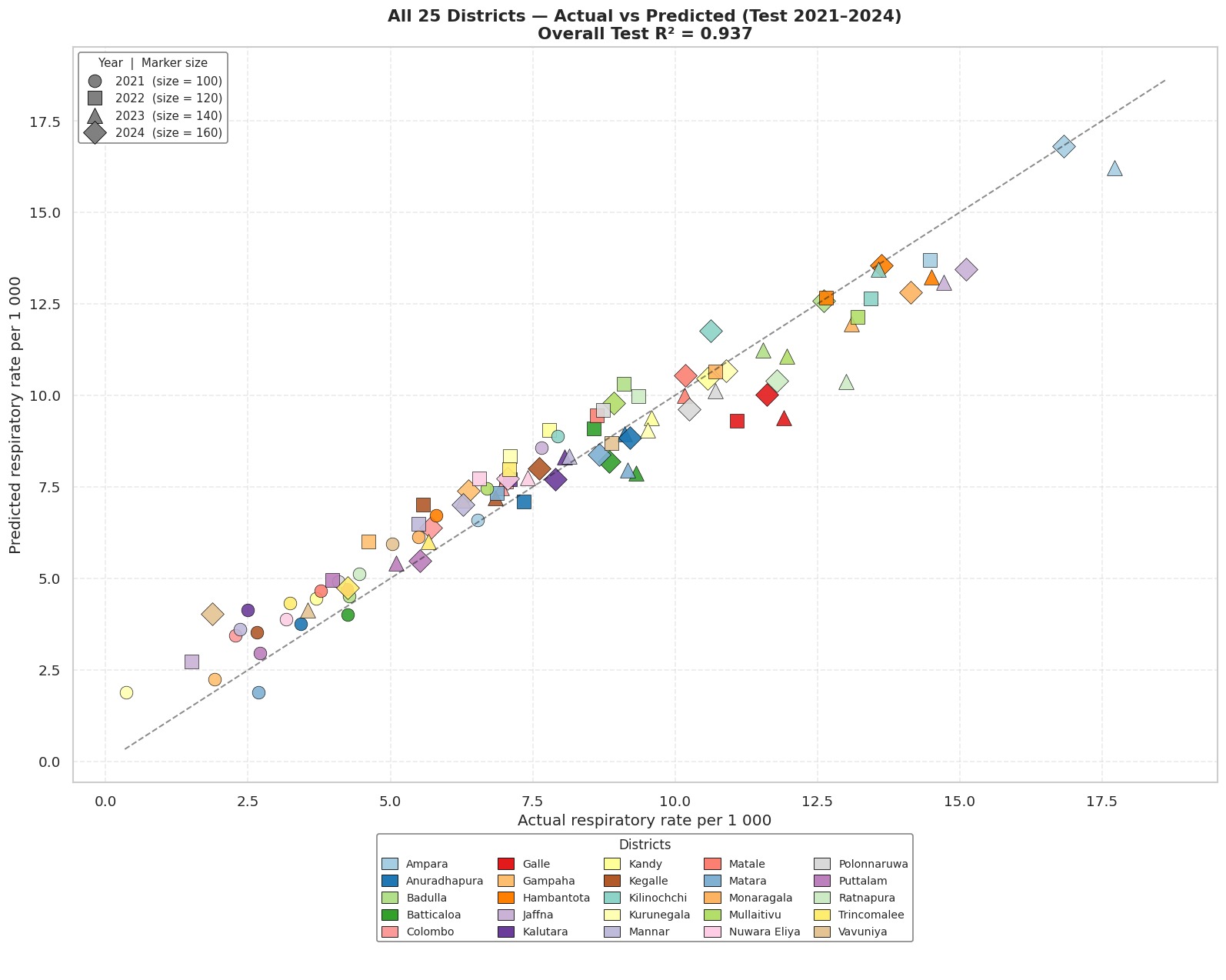}}
\caption{All 25 districts: actual vs.\ predicted total respiratory 
rate per 1{,}000 (test period 2021-2024). Each color represents 
a district. Overall test $R^2 = 0.937$.}
\label{fig:actual_pred}
\end{figure*}

\textit{1) Yearly Respiratory Rate Model (Main):} The model achieved a test $R^2 = 0.937$ (cross-validation (CV): $0.797 \pm 0.05$) with a Mean Absolute Error (MAE)\,=$0.776$ cases per 1{,}000 population. 
Figure~\ref{fig:actual_pred} shows the strong overlap along the diagonal for all 25 districts over the 2021-2024 test period. 
Table~\ref{tab:district_perf} shows the per-district validation, confirming that 21 out of 25 districts satisfy MAPE\,$\leq 20\%$. The model trained on pre-pandemic data (2015-2020) could not explain the anomalous MAPE of 114.7\% in Kurunegala due to COVID-19 pandemic reporting disruptions in 2020-2021. 
This is treated as a data quality artifact and not a failure of the model.
The gap between the cross-validation score ($0.797$) and the test
score ($0.937$) warrants caution: with only 150 training observations
and 49 features, the higher test value partly reflects comparatively
stable environmental and reporting conditions during 2021--2024 rather
than a uniform gain in generalization. Because every lagged and rolling
feature is computed in district-ordered fashion and the split is purely
temporal, look-ahead leakage is precluded as an explanation for the
high scores.

\begin{table}[htbp]
\caption{Per-District Prediction Quality (Test 2021-2024)}
\label{tab:district_perf}
\begin{center}
\begin{tabular}{lccc}
\hline
\textbf{District} & \textbf{MAE} & \textbf{MAPE (\%)} & 
\textbf{Quality} \\
\hline
Ampara       & 0.592 & 3.8   & Good \\
Anuradhapura & 0.276 & 4.7   & Good \\
Badulla      & 0.445 & 5.4   & Good \\
Hambantota   & 0.576 & 6.4   & Good \\
Kilinochchi  & 0.754 & 7.4   & Good \\
Kandy        & 0.581 & 10.0  & Good \\
Colombo      & 0.751 & 19.9  & Good \\
Kalutara     & 0.677 & 20.0  & Good \\
Mannar       & 0.795 & 21.3  & OK   \\
Jaffna       & 1.359 & 28.6  & OK   \\
Vavuniya     & 0.962 & 38.1  & Poor \\
Kurunegala   & 0.867 & 114.7 & Poor \\
\hline
\multicolumn{4}{l}{Good: MAPE\,$\leq 20\%$; OK: 20\%--35\%; 
Poor: $> 35\%$.}\\
\multicolumn{4}{l}{12 of 25 districts shown; remaining 13 all 
``Good''.}
\end{tabular}
\end{center}
\end{table}

\textit{2) Model of monthly PM\textsubscript{2.5}:} The model achieved a test $R^2$ of 0.976 (CV: 0.943), MAE of $0.520\,\mu$g/m$^3$, and RMSE of $0.717\,\mu$g/m$^3$, which is a significant improvement over a baseline model ($R^2 = 0.767$), defined as an XGBoost model trained on the pre-feature-engineering dataset---using only raw co-pollutant and meteorological inputs, without the engineered temporal lags, rolling averages, cross-domain interactions, or cyclic month encoding. The performance improvement is attributable to the incorporation of pollutant lags, rolling averages, cross-domain interactions, and cyclic month encoding.

\subsection{SHAP Feature Importance Analysis}
We calculated SHAP TreeExplainer values for the yearly respiratory model on 100 test observations (25 districts $\times$ 4 years), and we found four main results.

\begin{figure*}[htbp]
\centerline{\includegraphics[width=\linewidth]{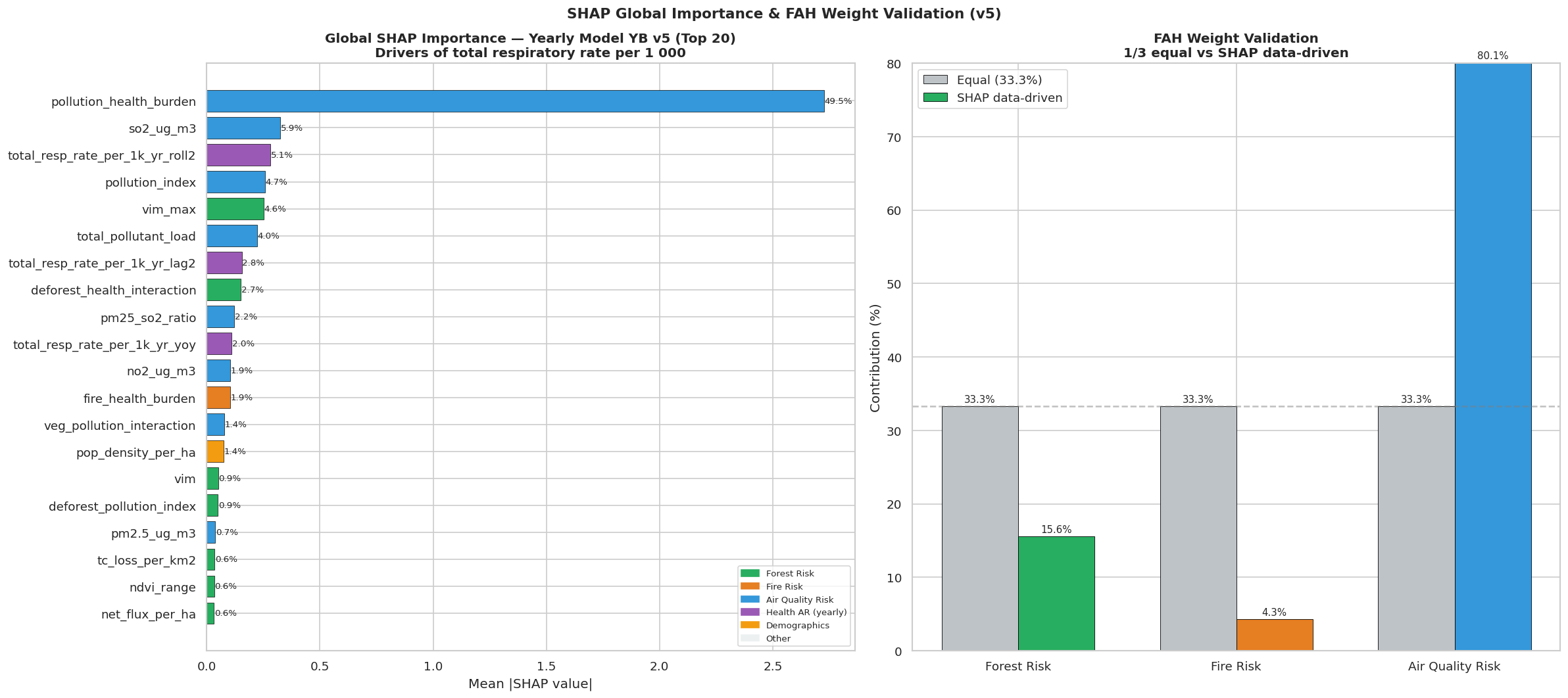}}
\caption{Left: Global SHAP importance for the yearly respiratory 
model (top 20 features), colour-coded by FAH component. Right: FAH 
weight validation comparing equal (33.3\%) weights against
SHAP-informed weights.}
\label{fig:shap}
\end{figure*}

\textit{1) Consolidated Global Feature Importance:} \texttt{pollution\_health\_burden} is the most important predictor (mean\,$|\text{SHAP}| = 2.73$, 49.5\% of total SHAP contribution), followed by SO\textsubscript{2} (5.9\%), the \texttt{yr\_roll2} respiratory lag features (5.1\%), the pollution index (4.7\%), and \texttt{vim\_max} (4.6\%), as shown in Fig.~\ref{fig:shap}.

\textit{2) Environmental SHAP Decomposition:} Grouping SHAP scores 
by environmental category reveals that Air Quality Risk attributes 
account for 80.1\% of the aggregate environmental SHAP signal, far 
exceeding Forest Risk (15.6\%) and Fire Risk (4.3\%). This 
decomposition contradicts naive equal-weight allocation of one third 
per domain and provides the empirical basis for the unequal FAH 
Risk Index weights.

\begin{figure}[htbp]
\centerline{\includegraphics[width=\columnwidth]{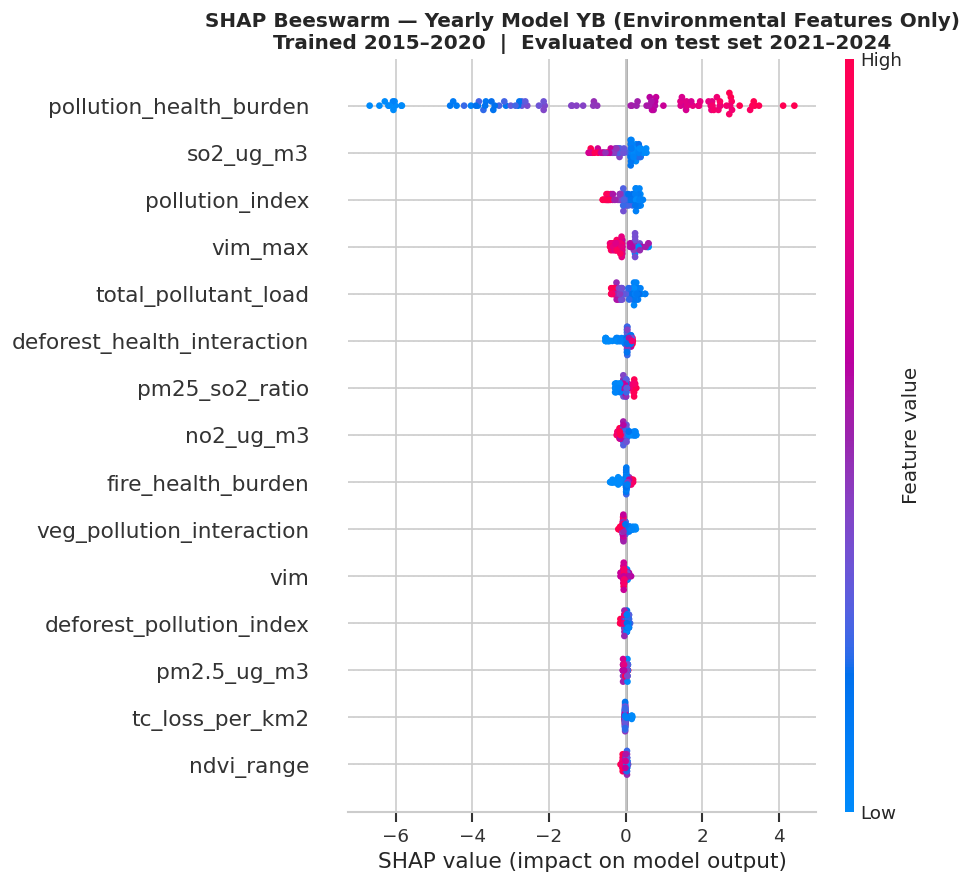}}
\caption{SHAP beeswarm plot for the yearly respiratory model 
(environmental features only). Red indicates high feature values; 
blue indicates low. Features are ordered by mean absolute SHAP 
contribution.}
\label{fig:beeswarm}
\end{figure}

\textit{3) SHAP Beeswarm Analysis:} The beeswarm plot 
(Fig.~\ref{fig:beeswarm}) confirms opposing directional effects: 
high values of \texttt{pollution\_health\_burden} increase predicted 
respiratory burden, whereas high values of 
\texttt{vim\_max}, indicative of healthy vegetation, exert a net 
protective effect, consistent with the inverse correlation observed 
in the EDA.

\begin{figure}[htbp]
\centerline{\includegraphics[width=\columnwidth]{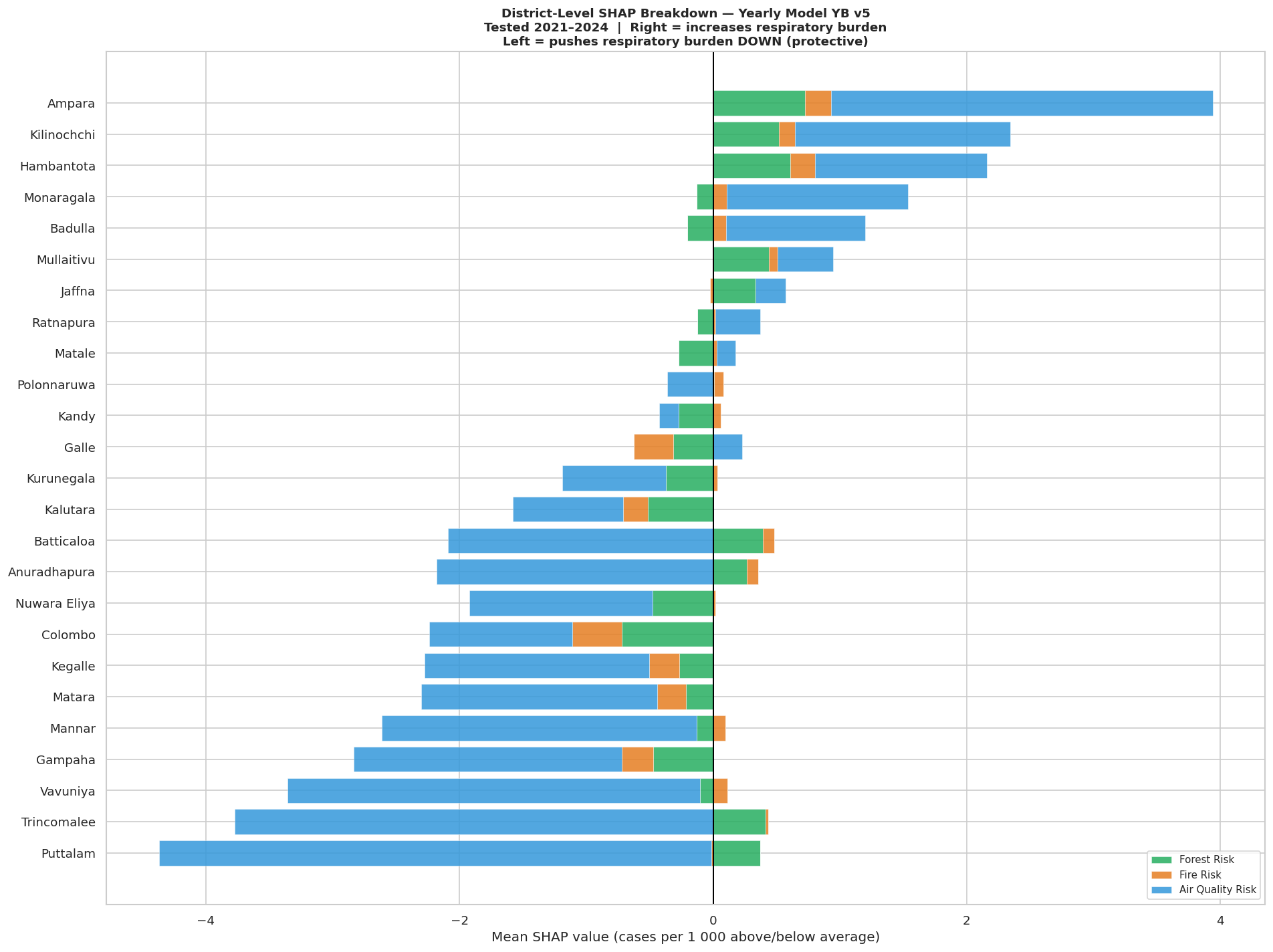}}
\caption{District-level SHAP breakdown by FAH component. Positive 
values (right) indicate features that increase predicted respiratory 
burden; negative values (left) indicate protective effects.}
\label{fig:district_shap}
\end{figure}

\textit{4) District-Level SHAP Decomposition:} 
Figure~\ref{fig:district_shap} presents the mean SHAP contribution 
of each FAH component by district. Western Province districts 
(Colombo, Gampaha, and Kalutara) exhibit the highest positive SHAP
scores, with respiratory risk attributed predominantly to the Air
Quality Risk component. Conversely, Eastern Province districts 
(Batticaloa and Trincomalee) show the most negative SHAP scores, 
reflecting a net protective effect against respiratory burden.

\subsection{Forest-Air-Health (FAH) Risk Index}
Using SHAP-derived weights (Forest: 15.6\%, Fire: 4.3\%, Air 
Quality: 80.1\%), each domain is represented by district-level mean 
sub-indicators individually normalized to $[0,1]$ using min-max 
normalization. Sub-indicators where higher values indicate lower 
risk (vegetation index median, VIM anomaly, and forest cover 
percentage) are inverted as $(1 - X_{norm})$ to ensure directional 
consistency. Normalized sub-indicators are averaged to produce a 
domain score, with five sub-indicators for the Forest and Fire 
domains and four for Air Quality (PM\textsubscript{2.5}, 
NO\textsubscript{2}, SO\textsubscript{2}, and composite pollution 
index). The final FAH score is computed as:

\begin{equation}
FAH = \lambda_{Forest} \cdot F_{forest} + 
      \lambda_{Fire} \cdot F_{fire} + 
      \lambda_{AQ} \cdot F_{air}
\end{equation}

where $\lambda_{Forest} = 0.156$, $\lambda_{Fire} = 0.043$, and
$\lambda_{AQ} = 0.801$. We stress that the FAH Risk Index is a
heuristic composite rather than a direct model output: its weights are
\emph{informed} by the SHAP importance decomposition of the
respiratory model, but the index itself is computed from normalized
environmental sub-indicators, not from model predictions. It should
therefore be interpreted as a transparent, SHAP-weighted
environmental-risk score rather than a direct estimate of respiratory
burden. Table~\ref{tab:fah} presents the top-5 and
bottom-5 districts. Colombo (0.802), Gampaha (0.708), and Kalutara 
(0.682) are the highest-risk districts, driven by elevated air 
quality scores. Kegalle (0.528), Ratnapura (0.470), Kurunegala 
(0.437), and Nuwara Eliya (0.425) fall in the moderate-risk tier, 
while the remaining 18 districts fall below the low-risk threshold 
($< 0.40$).

\begin{table}[htbp]
\caption{FAH Risk Index: District Rankings (Top 5 and Bottom 5)}
\label{tab:fah}
\begin{center}
\begin{tabular}{clcccc}
\hline
\textbf{Rank} & \textbf{District} & \textbf{FAH} & 
\textbf{Forest} & \textbf{Fire} & \textbf{Air} \\
\hline
1  & Colombo     & 0.802 & 0.595 & 0.018 & 0.884 \\
2  & Gampaha     & 0.708 & 0.448 & 0.286 & 0.781 \\
3  & Kalutara    & 0.682 & 0.523 & 0.143 & 0.742 \\
4  & Kegalle     & 0.528 & 0.400 & 0.102 & 0.576 \\
5  & Ratnapura   & 0.470 & 0.209 & 0.425 & 0.523 \\
\hline
21 & Polonnaruwa & 0.274 & 0.233 & 0.565 & 0.266 \\
22 & Vavuniya    & 0.265 & 0.427 & 0.549 & 0.218 \\
23 & Ampara      & 0.254 & 0.325 & 0.686 & 0.217 \\
24 & Trincomalee & 0.231 & 0.367 & 0.519 & 0.190 \\
25 & Batticaloa  & 0.208 & 0.345 & 0.508 & 0.165 \\
\hline
\end{tabular}
\end{center}
\end{table}

\begin{figure*}[htbp]
\centerline{\includegraphics[width=\linewidth]{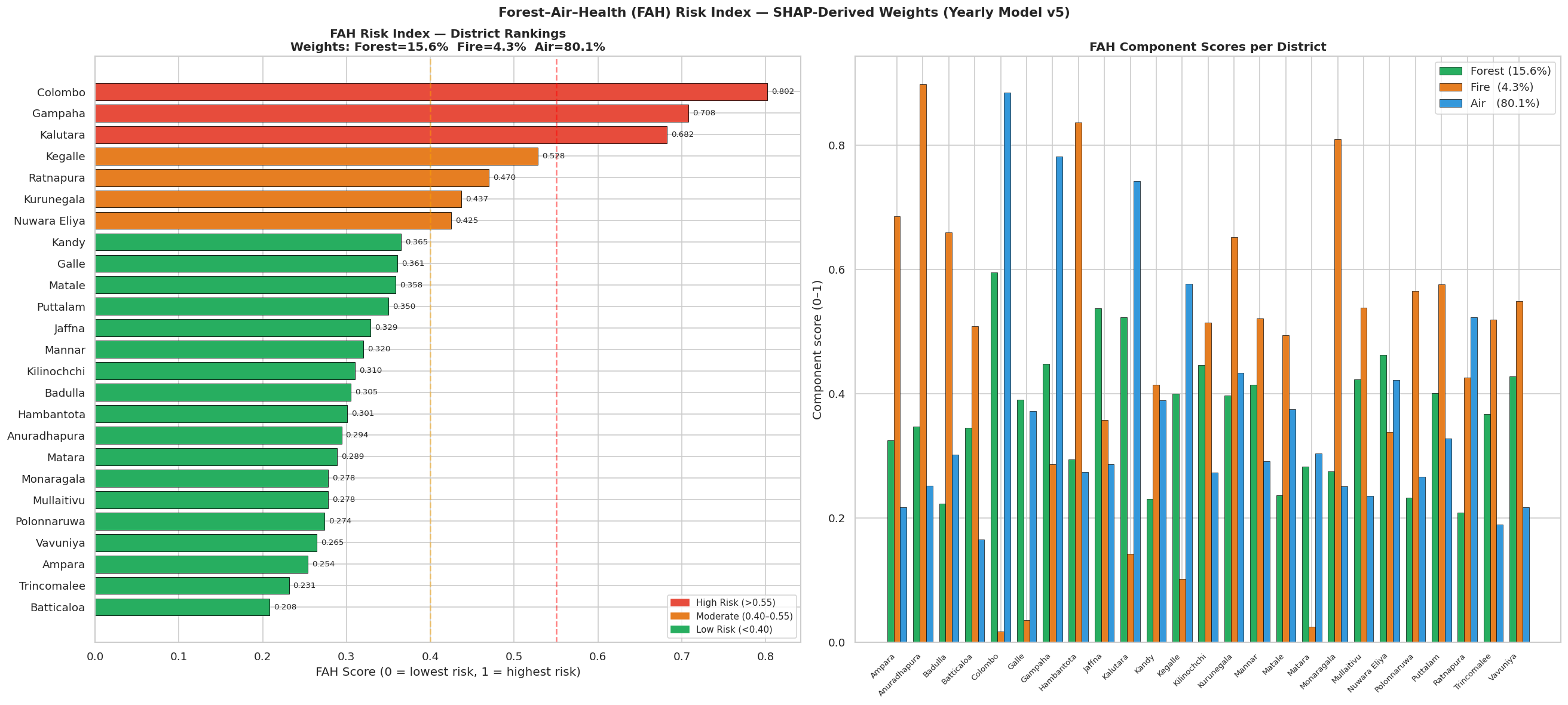}}
\caption{FAH Risk Index. Left: district rankings with risk tier 
classification (red\,$=$\,high, orange\,$=$\,moderate, 
green\,$=$\,low). Right: decomposed FAH component scores per 
district showing the dominance of air quality in driving composite 
risk.}
\label{fig:fah}
\end{figure*}

\subsection{Study Limitations}
Several limitations should be noted. Most importantly, this study
characterizes \emph{associations} rather than causal effects, and the
models do not include healthcare access, which our own results
identify as a key confounder: districts with comparable environmental
exposure but stronger healthcare infrastructure (e.g., Colombo vs.\
Jaffna) report markedly different respiratory rates. District-level
proxies for healthcare access (e.g., hospital beds or physicians per
capita), together with demographic, smoking-prevalence, urbanization,
and socioeconomic covariates, were unavailable in harmonized 11-year
form and are left to future work; their omission means the reported
SHAP rankings should be read as the strongest predictors \emph{within
the available environmental feature set} rather than as the dominant
determinants of respiratory burden overall. Second, health data were
annualized uniformly across months, which discards within-year
seasonality (e.g., monsoon-linked spikes) and weakens temporal
alignment between predictors and outcomes; the yearly respiratory
model was adopted specifically to limit this effect, and monthly
modeling is therefore restricted to PM\textsubscript{2.5}. Third, the
coarse resolution of MERRA-2 ($0.5^\circ \times 0.625^\circ$) and CAMS
grids may conceal sub-district pollution sources, and the
PM\textsubscript{2.5} model lacks an explicit trans-boundary transport
indicator (e.g., wind direction, Indian AQI, or neighboring-region
biomass-burning indices), even though the MERRA-2 inputs implicitly
capture some cross-border smog from Indian agricultural burning; the
residual PM\textsubscript{2.5} variance ($\sim$2.4\%) likely reflects
these omissions alongside microclimate and household-cooking
emissions. Finally, accuracy is weakest in a few districts (Vavuniya,
Jaffna, and especially Kurunegala, MAPE\,$= 114.7\%$), where COVID-19
reporting disruptions (2020--2021) distorted admission records; these
are treated as data-quality artifacts rather than model failures.

%----------------------------------------------------------------------
\section{Conclusion and Future Work}
%----------------------------------------------------------------------

This is the first nationwide district-level study in Sri Lanka to examine 
the relationships between forest degradation, atmospheric pollution, and 
respiratory health over the period 2014-2024. Five principal findings 
were established: (1) forest degradation is associated with reduced
natural particulate filtration capacity of vegetation ($r = -0.40$
between vegetation index and PM\textsubscript{2.5});
(2) air quality is the strongest predictor of respiratory rate in the
model, comprising 80.1\% of the aggregate environmental SHAP signal,
ahead of forest degradation (15.6\%) and fire activity (4.3\%), though
this attribution reflects predictive importance rather than
established causation; (3) temporally 
cross-validated XGBoost models achieved high predictive accuracy for 
yearly respiratory rate ($R^2 = 0.937$) and monthly PM\textsubscript{2.5} 
($R^2 = 0.976$); (4) the FAH Risk Index identifies Colombo, Gampaha, 
and Kalutara as the highest-risk districts; and (5) PCA and $K$-means 
clustering revealed six environmental risk axes and four distinct district 
archetypes. Future work will focus on incorporating monthly health records 
and vehicle emission data, developing Auto-regressive Integrated Moving Average (ARIMA) and Long Short-Term Memory (LSTM) forecasting models to project district-level PM\textsubscript{2.5} concentrations to 2030, and operationalizing the FAH Risk Index as a deployable early-warning system for public health agencies.

% %----------------------------------------------------------------------
% \section*{Acknowledgment}
% %----------------------------------------------------------------------

% The authors thank the following data providers: Global Forest Watch (World Resources Institute), NASA Giovanni and MERRA-2 (NASA GES DISC), Copernicus Atmosphere Monitoring Service (ECMWF), GLAM/VIIRS vegetation index service (NASA GSFC), NASA FIRMS active fire data (SUOMI VIIRS C2), Department of Census and Statistics Sri Lanka, and the Ministry of Health Sri Lanka. District border information was gathered from the Humanitarian Data Exchange (HDX). We are grateful to the Department of Computer Science and Engineering at the University of Moratuwa for institutional assistance.

% %----------------------------------------------------------------------

{\footnotesize
\bibliographystyle{IEEEtranN}
\bibliography{references}
}

\end{document}